\documentclass[sigplan]{acmart}
\renewcommand\footnotetextcopyrightpermission[1]{}

\usepackage{kotex}
\usepackage{subcaption} 
\usepackage{hhline}
\usepackage[table]{xcolor}
\usepackage{graphicx}
\usepackage{pifont}
\usepackage{xcolor}

\usepackage{wrapfig}
\usepackage{enumitem} 

\AtBeginDocument{%
  }

\acmConference[SAA'25]{1st Workshop on Systems for Agentic AI}{October 13, 2025}{Seoul, Korea}

\begin{document}

\title{GPU Memory Prediction for Multimodal Model Training} 

\author{Jinwoo Jeong}
\authornote{The first two authors contributed equally to this work.}
\affiliation{%
  \institution{Korea University}
  \city{Seoul}
  \country{Republic of Korea}
}
\email{jwjeong@os.korea.ac.kr}

\author{Minchul Kang}
\authornotemark[1]
\affiliation{%
  \institution{Korea University}
  \city{Seoul}
  \country{Republic of Korea}
}
\email{mckang@os.korea.ac.kr}

\author{Younghun Go}
\affiliation{%
  \institution{Korea University}
  \city{Seoul}
  \country{Republic of Korea}
}
\email{yhgo@os.korea.ac.kr}

\author{Changyong Shin}
\affiliation{%
  \institution{Korea University}
  \city{Seoul}
  \country{Republic of Korea}
}
\email{cyshin@os.korea.ac.kr}

\author{Hyunho Lee}
\affiliation{%
  \institution{Korea University}
  \city{Seoul}
  \country{Republic of Korea}
}
\email{hhlee@os.korea.ac.kr}

\author{Junho Yoon}
\affiliation{%
  \institution{KT Corporation}
  \city{Seoul}
  \country{Republic of Korea}
}
\email{jun\_ho.yoon@kt.com}

\author{Gyeongsik Yang}
\affiliation{%
  \institution{Korea University}
  \city{Seoul}
  \country{Republic of Korea}
}
\email{g\_yang@korea.ac.kr}

\author{Chuck Yoo}
\affiliation{%
  \institution{Korea University}
  \city{Seoul}
  \country{Republic of Korea}
}
\email{chuckyoo@os.korea.ac.kr}

\renewcommand{\shortauthors}{Jeong et al.}

\begin{abstract}

As deep learning models in agentic AI systems grow in scale and complexity, GPU memory requirements increase and often exceed the available GPU memory capacity, so that out-of-memory (OoM) errors occur. It is well known that OoM interrupts the whole training itself and wastes substantial computational resources. Therefore, to prevent OoM, accurate prediction of GPU memory usage is essential. However, previous studies focus only on unimodal architectures and fail to generalize to multimodal models, even though the multimodal models are a common choice in agentic AI systems. To address this limitation, we propose a framework that predicts the peak GPU memory usage by analyzing the model architecture and training behavior of multimodal models. Specifically, the framework decomposes the multimodal model into its constituent layers and applies “factorization” to estimate the memory usage of each layer. Our evaluation shows that our framework achieves high prediction accuracy of $\sim$8.7\% average MAPE.

\end{abstract}
\maketitle
\section{Introduction}

Agentic AI systems increasingly rely on multimodal models to perform complex tasks. In particular, vision–language models, a representative type of multimodal architectures, serve as a core component of Agentic AI systems because they can integrate diverse modalities and adapt to tool interfaces \cite{qian}. For example, to enable an agent to plan actions with the external tools (e.g., databases, search engines, or code execution) in response to user queries, the models are trained or fine-tuned to generate API calls that understand such tools \cite{gao2024multi}.
Recently, the increasing scale and complexity of the models have sharply raised GPU memory requirements during training. When the peak GPU memory usage consumed by the model exceeds the available capacity, OoM errors occur, interrupting training and often wasting substantial computational resources. Therefore, accurately predicting GPU memory usage in advance is critical to avoid OoM errors and ensure efficient, stable training.

To address this need, several studies have proposed methods for predicting GPU memory usage in unimodal architectures: 1) profiling-based prediction \cite{xonar, driple, dnnmem}, which runs a few training iterations to predict peak memory usage---assuming that training consists of repeated, identical forward and backward propagation steps, and 2) formulation-based modeling \cite{accelerating, llmem}, which predicts the memory usage of individual layers in the model architecture to determine the GPU memory consumption.

However, recently multimodal models pose significant challenges for memory prediction. Such models comprise different types of modules (e.g., vision encoders and language decoders), each containing various layers (e.g., embedding and feed-forward networks). For example, the representative multimodal model LLaVA \cite{llava} consists of several hundred layers across multiple modules, making memory usage prediction highly challenging. Furthermore, the memory usage of a layer is determined by several factors, such as parameters and gradients. Thus, existing prediction methods have the following limitations. First, existing profiling-based methods require multiple pre-training runs, causing significant overhead that can take a long time \cite{xonar}. Second, formulation-based methods work on specific layer types or particular model architectures so that they have to come up with a new formula each time when new modules are added to multimodal models. We have applied the method in \cite{accelerating} to a multimodal model, but we found that it does not work at all because the formula was designed for a specific unimodal architecture and does not generalize to heterogeneous modules in multimodal models. To our knowledge, there is no literature that applies the profiling method to a multimodal model.

To address this limitation, we propose a framework that accurately predicts the peak GPU memory usage by analyzing the architecture and training behavior of multimodal models. Our evaluation on the LLaVA-1.5 (7B) model demonstrates that, even under diverse hyperparameter settings, our framework achieves MAPE as low as 8.7\%.

\section{Background \& Motivation}
\label{sec:motiv}
\vspace{0.075in}\noindent
\textbf{Multimodal models.}
Multimodal models serve as a core component in Agent AI systems to plan the actions with various agent tools for the given request. It uses heterogeneous modules and layers to handle the tasks of different natures such as image captioning and image-text reasoning. This heterogeneity makes variations in the types and configurations of layers within each module. A representative model is LLaVA \cite{llava} that uses the pre-trained CLIP ViT-L/14 \cite{clip} as a vision encoder module to extract image features, and a Vicuna-based \cite{vicuna} model as a language decoder module to generate text. These two are connected by a projection layer that aligns the vision module’s output with the language module's input.

Training of LLaVA proceeds in two stages: 1) pre-training, 2) fine-tuning. In the pre-training stage, only the projection layer is updated, while the vision module and language module remain frozen. In the subsequent fine-tuning stage, the projection layer and parts of the language module are updated, while the vision module remains frozen \cite{llava}. Note that this training behavior may vary depending on multimodal models and fine-tuning techniques (e.g., LoRA \cite{lora}). Such training behavior inevitably changes the GPU memory usage, in contrast to unimodal models. So it is very challenging to predict GPU memory usage for multimodal models.

\vspace{0.075in}\noindent
\textbf{GPU memory usage.}
In training DL models, the peak GPU memory usage is affected by multiple factors. The main factors are:
\begin{itemize}[leftmargin=10pt, itemsep=0pt]  
    \item \textbf{Model parameters ($M_\text{param}$)}: 
    weights and biases for each layer that must remain resident in GPU memory throughout forward and backward passes, and their memory footprint scales with model size.
    \item \textbf{Optimizer states ($M_\text{opt}$)}: additional parameter states maintained by the optimizer to facilitate parameter updates, such as momentum terms and variance estimates in Adam.
    \item \textbf{Gradients ($M_\text{grad}$)}: partial derivatives of the loss for each parameter, computed during backpropagation. They are stored temporarily until the optimizer step is performed.
    \item \textbf{Activations ($M_\text{act}$)}: intermediate outputs of each layer in the forward pass. They must be stored in memory until their corresponding backward computations are complete.
\end{itemize}

Previous studies based on formula \cite{accelerating, llmem} predict memory usage from these factors, but their methods focused on unimodal architectures. Multimodal models differ in having heterogeneous modules with distinct training behaviors. This motivates our framework that predicts GPU memory usage by explicitly modeling the architecture and training behavior of multimodal models.

\begin{figure}[t]
    \centering 
    \includegraphics[width=\linewidth]{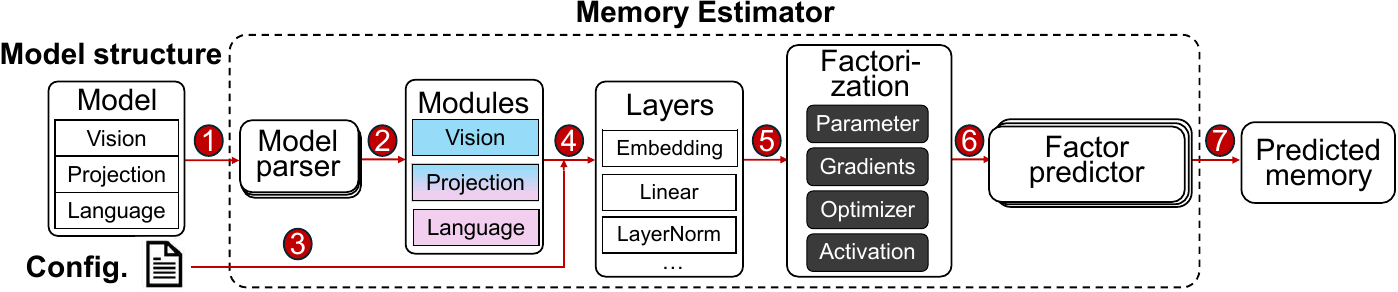}
    \caption{Workflow of the proposed framework.}
    \label{fig:1}
    \hfill
\end{figure}

\section{Design}

Fig.~\ref{fig:1} shows the workflow of the proposed framework. It begins with the \textit{Model parser} (\ding{172}) that analyzes the model architecture and extracts key ``modules'' (\ding{173}) based on modality. A configuration file provides training hyperparameters such as batch size (\ding{174}). Each parsed module is then further decomposed into fine-grained ``layers'', such as \texttt{nn.Linear} (\ding{175}), to facilitate the next step called “factorization”. Specifically, we identify the layers within the model's modules using the PyTorch API. The framework factorizes memory usage for each layer into four factors: model parameters, gradients, optimizer states, and activations (\ding{176}). Note that each layer can have different factors. For example, an embedding layer in a frozen vision module has neither gradients nor optimizer states, whereas a feed-forward layer in a language module requires both in addition to its parameters. Thus, this factorization reflects the structure and training behavior of multimodal models.

We then predict the peak memory usage of each layer by applying a per-factor analytical equation, called \textit{factor predictor} (\ding{177}). In this paper, we focus on presenting the overall framework and do not include detailed derivations of each factor. For each layer, the factor predictor computes four factors (e.g., parameters and gradients), and it repeats the calculation for all layers across all modules. 

Formally, we predict the peak memory usage as follows:
\begin{equation}
M_{\text{peak}} = \sum_{\text{module}} \sum_{\text{layer}}
\left( M_{\text{param}} + M_{\text{opt}} + M_{\text{grad}} + M_{\text{act}} \right)
\label{eq:memory_estimate}
\end{equation}
Here, $M_{act}$ in a multimodal model is obtained by computing activations for modalities whose parameters are being updated, whereas in a unimodal model it is computed for the single modality being trained. Finally, the framework generates the predicted peak memory usage (\ding{178}), which is used to prevent OoM errors.

\section{Evaluation}

\begin{figure}[t]
    \centering
    \begin{subfigure}[t]{0.44\linewidth} 
        \centering
        \includegraphics[width=\linewidth]{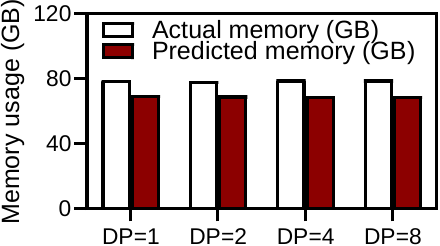}
        \caption{SeqLen=1024, MBS=16.}
        \label{fig:2}
    \end{subfigure}
    \hfill
    \begin{subfigure}[t]{0.44\linewidth}
        \centering
        \includegraphics[width=\linewidth]{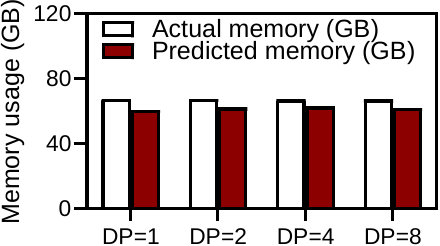}
        \caption{SeqLen=2048, MBS=8.}
        \label{fig:3}
    \end{subfigure}
    \caption{GPU memory usage prediction results.} 
    \label{fig:side-by-side}
\end{figure}

To demonstrate that our framework maintains robust predictive accuracy under varying hyperparameter configurations, we evaluate the LLaVA-1.5 (7B) model in two hyperparameter settings. In the first setting, we set the sequence length (SeqLen) to 1,024 and the micro-batch size (MBS) to 16, varying the data parallelism (DP) from 1 to 8.
In the second setting, we use SeqLen of 2048 and MBS of 8, again varying DP from 1 to 8. All experiments run on a single GPU node with eight NVIDIA H100 80GB GPUs (NVLink), using PyTorch 24.07~\cite{pytorch} and ZeRO-2 from DeepSpeed~\cite{deepspeed}.

Fig.~\ref{fig:2} shows the prediction accuracy for the first setting. We observe an average MAPE of 13\% across various DP degrees. Fig.~\ref{fig:3} shows the prediction accuracy for the second setting. The results show that the average MAPE is 8.7\%, demonstrating that our framework provides consistent and reliable prediction accuracy across various training setups.

\section{Future work}

In the future, we plan to extend our memory usage prediction to include other optimization techniques such as parameter-efficient fine-tuning and kernel fusion. We also plan to extend our memory prediction to inference workloads of agentic AI systems that manage memory with key-value caching and multi-turn orchestration.

\section{Conclusion}
In this paper, we propose a framework to predict GPU memory usage for training multimodal models. The framework parses the model to extract modules. Each module is further decomposed into individual layers, and memory usage is predicted based on four factors. The factor predictor applies per-factor analytical equations to each layer, and the predictions are aggregated to compute the peak GPU memory usage. The proposed framework achieves high prediction accuracy, with MAPE values of $\sim$8.7\% in two different settings, demonstrating reliable and consistent performance across various hyperparameters.

\begin{acks}
This work was supported by KT (Korea Telecom)-Korea University AICT R\&D Center. Corresponding authors are Gyeongsik Yang and Chuck Yoo.
\end{acks}

\bibliographystyle{ACM-Reference-Format}
\bibliography{samples/references}

@misc{llmem,
      title={LLMem: Estimating GPU Memory Usage for Fine-Tuning Pre-Trained LLMs}, 
      author={Taeho Kim and Yanming Wang and Vatshank Chaturvedi and Lokesh Gupta and Seyeon Kim and Yongin Kwon and Sangtae Ha},
      year={2024},
      eprint={2404.10933},
      archivePrefix={arXiv},
      primaryClass={cs.AI},
      url={https://arxiv.org/abs/2404.10933}, 
}

@inproceedings{dnnmem,
  title={Estimating GPU memory consumption of deep learning models},
  author={Gao, Yanjie and Liu, Yu and Zhang, Hongyu and Li, Zhengxian and Zhu, Yonghao and Lin, Haoxiang and Yang, Mao},
  booktitle={Proceedings of the 28th ACM Joint Meeting on European Software Engineering Conference and Symposium on the Foundations of Software Engineering},
  pages={1342--1352},
  year={2020}
}

@article{accelerating,
  title={Accelerating large language model training with 4d parallelism and memory consumption estimator},
  author={Fujii, Kazuki and Watanabe, Kohei and Yokota, Rio},
  journal={arXiv preprint arXiv:2411.06465},
  year={2024}
}

@article{llava,
  title={Visual instruction tuning},
  author={Liu, Haotian and Li, Chunyuan and Wu, Qingyang and Lee, Yong Jae},
  journal={Advances in neural information processing systems},
  volume={36},
  pages={34892--34916},
  year={2023}
}

@article{pytorch,
  title={Pytorch: An imperative style, high-performance deep learning library},
  author={Paszke, Adam and Gross, Sam and Massa, Francisco and Lerer, Adam and Bradbury, James and Chanan, Gregory and Killeen, Trevor and Lin, Zeming and Gimelshein, Natalia and Antiga, Luca and others},
  journal={Advances in neural information processing systems},
  volume={32},
  year={2019}
}

@inproceedings{deepspeed,
  title={Deepspeed: System optimizations enable training deep learning models with over 100 billion parameters},
  author={Rasley, Jeff and Rajbhandari, Samyam and Ruwase, Olatunji and He, Yuxiong},
  booktitle={Proceedings of the 26th ACM SIGKDD international conference on knowledge discovery \& data mining},
  pages={3505--3506},
  year={2020}
}

@inproceedings{clip,
  title={Learning transferable visual models from natural language supervision},
  author={Radford, Alec and Kim, Jong Wook and Hallacy, Chris and Ramesh, Aditya and Goh, Gabriel and Agarwal, Sandhini and Sastry, Girish and Askell, Amanda and Mishkin, Pamela and Clark, Jack and others},
  booktitle={International conference on machine learning},
  pages={8748--8763},
  year={2021},
  organization={PmLR}
}

@article{vicuna,
  title={Vicuna: An open-source chatbot impressing gpt-4 with 90\%* chatgpt quality},
  author={Chiang, Wei-Lin and Li, Zhuohan and Lin, Ziqing and Sheng, Ying and Wu, Zhanghao and Zhang, Hao and Zheng, Lianmin and Zhuang, Siyuan and Zhuang, Yonghao and Gonzalez, Joseph E and others},
  journal={See https://vicuna. lmsys. org (accessed 14 April 2023)},
  volume={2},
  number={3},
  pages={6},
  year={2023}
}

@inproceedings{xonar,
  title={Xonar: Profiling-based job orderer for distributed deep learning},
  author={Shin, Changyong and Yang, Gyeongsik and Yoo, Yeonho and Lee, Jeunghwan and Yoo, Chuck},
  booktitle={2022 IEEE 15th International Conference on Cloud Computing (CLOUD)},
  pages={112--114},
  year={2022},
  organization={IEEE}
}

@article{driple,
  title={Prediction of the resource consumption of distributed deep learning systems},
  author={Yang, Gyeongsik and Shin, Changyong and Lee, Jeunghwan and Yoo, Yeonho and Yoo, Chuck},
  journal={Proceedings of the ACM on Measurement and Analysis of Computing Systems},
  volume={6},
  number={2},
  pages={1--25},
  year={2022},
  publisher={ACM New York, NY, USA}
}

@article{lora,
  title={Lora: Low-rank adaptation of large language models.},
  author={Hu, Edward J and Shen, Yelong and Wallis, Phillip and Allen-Zhu, Zeyuan and Li, Yuanzhi and Wang, Shean and Wang, Lu and Chen, Weizhu and others},
  journal={ICLR},
  volume={1},
  number={2},
  pages={3},
  year={2022}
}

@article{gao2024multi,
  title={Multi-modal agent tuning: Building a vlm-driven agent for efficient tool usage},
  author={Gao, Zhi and Zhang, Bofei and Li, Pengxiang and Ma, Xiaojian and Yuan, Tao and Fan, Yue and Wu, Yuwei and Jia, Yunde and Zhu, Song-Chun and Li, Qing},
  journal={arXiv preprint arXiv:2412.15606},
  year={2024}
}

@article{qian,
  title={Agentthink: A unified framework for tool-augmented chain-of-thought reasoning in vision-language models for autonomous driving},
  author={Qian, Kangan and Jiang, Sicong and Zhong, Yang and Luo, Ziang and Huang, Zilin and Zhu, Tianze and Jiang, Kun and Yang, Mengmeng and Fu, Zheng and Miao, Jinyu and others},
  journal={arXiv preprint arXiv:2505.15298},
  year={2025}
}

@String{Computing = "Computing" }
\end{document}